\documentclass{article}
\usepackage{cite}
\usepackage{amsmath,graphicx,mlspconf}
\usepackage{amsmath}
\usepackage{amsmath,amssymb,amsfonts}

\usepackage{array}
\usepackage{float}
\usepackage[caption=false,font=footnotesize]{subfig}
\usepackage{stfloats}
\usepackage{url}
\usepackage{makecell}
\usepackage[caption=false, font=footnotesize]{subfig}
\newcolumntype{P}[1]{>{\centering\arraybackslash}p{#1}}
\usepackage{algorithm}
\usepackage{algpseudocode}

\renewcommand{\algorithmicensure}{\textbf{Output:}}
\usepackage{xcolor}
\DeclareMathOperator{\SNR}{SNR}
\DeclareMathOperator{\NMSE}{NMSE}
\DeclareMathOperator{\softmax}{softmax}

\newcommand{\bmat}{\begin{bmatrix}}
\newcommand{\emat}{\end{bmatrix}}

\newcommand{\ones}{\mathbf 1}

\newcommand{\vect}[1]{\boldsymbol{#1}}

\newcommand{\norm}[1]{\lVert #1 \rVert}

\DeclareMathOperator*{\He}{H}
\DeclareMathOperator*{\Tr}{T}

\newcommand{\vQ}{{\vect Q}}

\newcommand{\vr}{{\vect r}}

\newcommand{\vh}{{\vect h}}

\newcommand{\vb}{{\vect b}}
\newcommand{\vc}{{\vect c}}

\newcommand{\vn}{{\vect n}}
\newcommand{\vy}{{\vect y}}

\newcommand{\vz}{{\vect z}}

\newcommand{\vf}{{\vect f}}
\newcommand{\vu}{{\vect u}}
\newcommand{\vo}{{\vect o}}
\newcommand{\vi}{{\vect i}}

\newcommand{\vK}{{\vect K}}
\newcommand{\vY}{{\vect Y}}
\newcommand{\vX}{{\vect X}}

\newcommand{\vH}{{\vect H}}
\newcommand{\vZ}{{\vect Z}}

\newcommand{\vV}{{\vect V}}
\newcommand{\vW}{{\vect W}}
\newcommand{\vS}{{\vect S}}

\usepackage{tikz}
\usetikzlibrary{fit}
\usepackage{ellipsis}
\usetikzlibrary{calc}
\usetikzlibrary{decorations.pathreplacing,decorations.markings,shapes.geometric}
\usetikzlibrary{chains}
\tikzstyle{block} = [draw, rectangle, minimum height=3em, minimum width=4.5em]
\tikzstyle{input} = [coordinate]
\tikzstyle{output} = [coordinate]
\tikzstyle{pinstyle} = [pin edge={to-,thin,black}]

\usetikzlibrary{positioning}

\tikzset{radiation/.style={{decorate,decoration={expanding waves,angle=90,segment length=5pt}}}}
\tikzset{font={\fontsize{10pt}{12}\selectfont}}
\usepackage[left=1.62cm,right=1.62cm,top=1.9cm]{geometry}

\usetikzlibrary{spy}
\usepackage{pgfplots}
\usepackage{wrapfig}
\usepackage[normalem]{ulem}
\usetikzlibrary{arrows,shapes}
\usetikzlibrary{positioning,shapes.callouts}
\usepgfplotslibrary{groupplots,dateplot}
\usetikzlibrary{patterns,shapes.arrows}
\pgfplotsset{compat=newest}

\usepackage[acronym,shortcuts]{glossaries}
\usepackage[acronym,shortcuts]{glossaries}
\newacronym{bs}{BS}{base station}
\newacronym{csi}{CSI}{channel state information}
\newacronym{fdd}{FDD}{frequency division duplex}
\newacronym{mimo}{MIMO}{multiple-input multiple-output}
\newacronym{nn}{NN}{neural network}
\newacronym{mt}{MT}{mobile terminal}
\newacronym{pca}{PCA}{principal component analysis}
\newacronym{ai}{AI}{artificial intelligence}
\newacronym{dl}{DL}{downlink}
\newacronym{ul}{UL}{uplink}
\newacronym{ae}{AE}{autoencoder}
\newacronym{ar}{AR}{autoregressive}
\newacronym{kf}{KF}{Kalman filter}
\newacronym{rnn}{RNN}{recurrent neural network}
\newacronym{mar}{MAR}{multivariate autoregressive}
\newacronym{lstm}{LSTM}{long-short-term memory}
\newacronym{seqa}{Seq2Seq-attn}{sequence-to-sequence with attention}
\newacronym{seq}{Seq2Seq}{sequence-to-sequence}
\newacronym{gru}{GRU}{gated recurrent unit}
\newacronym{mlp}{MLP}{multi-layer perceptron}
\newacronym{cnn}{CNN}{convolutional neural network}
\newacronym{ln}{LN}{layer normalization}
\newacronym{mse}{MSE}{mean squared error}
\newacronym{nmse}{$\NMSE$}{normalized mean squared error}
\newacronym{pe}{PE}{positional encoding}
\newacronym{rpe}{RPE}{reversed positional encoding}
\newacronym{flop}{FLOP}{floating point operations}
\definecolor{tum-pink}{HTML}{B55CA5}
\definecolor{tum-pink-dark}{HTML}{9B468D}
\definecolor{tum-pink-1}{HTML}{C680BB}
\definecolor{tum-pink-2}{HTML}{D6A4CE}
\definecolor{tum-pink-3}{HTML}{E6C7E1}
\definecolor{tum-pink-4}{HTML}{F6EAF4}

\definecolor{tum-blue-dark-5}{HTML}{165DB1}
\definecolor{tum-blue-light}{HTML}{5E94D4}
\definecolor{tum-blue-light-dark}{HTML}{9ABCE4}
\definecolor{tum-blue-light-2}{HTML}{C2D7EF}
\definecolor{tum-blue-light-3}{HTML}{D7E4F4}
\definecolor{tum-blue-light-4}{HTML}{E3EEFA}

\definecolor{tum-yellow-dark}{HTML}{CBAB01}
\definecolor{tum-yellow}{HTML}{FED702}
\definecolor{tum-yellow-1}{HTML}{FEDE34}
\definecolor{tum-yellow-2}{HTML}{FEE667}
\definecolor{tum-yellow-3}{HTML}{FEEE9A}
\definecolor{tum-yellow-4}{HTML}{FEF6CD}

\definecolor{tum-orange}{HTML}{F7811E}
\definecolor{tum-orange-dark}{HTML}{D99208}
\definecolor{tum-orange-1}{HTML}{F9BF4E}
\definecolor{tum-orange-2}{HTML}{FAD080}
\definecolor{tum-orange-3}{HTML}{FCE2B0}

\definecolor{tum-red-dark}{HTML}{D95117}
\definecolor{tum-red-1}{HTML}{EF9067}	
\definecolor{tum-red-2}{HTML}{F3B295}	
\definecolor{tum-red-3}{HTML}{F6C2AC}
\definecolor{tum-red-4}{HTML}{FBEADA}

\definecolor{tum-blue-bright}{HTML}{8F81EA}
\definecolor{tum-blue-bright-dark}{HTML}{6955E2}
\definecolor{tum-blue-bright-1}{HTML}{B6ACF1}
\definecolor{tum-blue-bright-2}{HTML}{C9C2F5}	
\definecolor{tum-blue-bright-3}{HTML}{DCD8F9}
\definecolor{tum-blue-bright-4}{HTML}{EFEDFC}

\definecolor{tum-green}{HTML}{9FBA36}	
\definecolor{tum-green-dark}{HTML}{7D922A}	
\definecolor{tum-green-1}{HTML}{B6CE55}
\definecolor{tum-green-2}{HTML}{C7D97D}	
\definecolor{tum-green-3}{HTML}{D8E5A4}	
\definecolor{tum-green-4}{HTML}{E9F1CB}

\definecolor{tum-grey-1}{HTML}{20252A}
\definecolor{tum-grey-2}{HTML}{333A41}	
\definecolor{tum-grey-3}{HTML}{475058}	
\definecolor{tum-grey-4}{HTML}{6A757E}	
\definecolor{tum-grey-7}{HTML}{DDE2E6}

\definecolor{crimson2143940}{RGB}{214,39,40}
\definecolor{darkgray176}{RGB}{176,176,176}
\definecolor{darkorange25512714}{RGB}{255,127,14}
\definecolor{forestgreen4416044}{RGB}{44,160,44}
\definecolor{lightgray204}{RGB}{204,204,204}
\definecolor{steelblue31119180}{RGB}{31,119,180}

\RequirePackage{xcolor}
\RequirePackage{kvoptions}%
\SetupKeyvalOptions{family=tumcolor,prefix=tumcolor@}%
\DeclareBoolOption{print}%
\ProcessKeyvalOptions*\relax

    \definecolor{TUMBlack}           {cmyk}{0.00,0.00,0.00,1.00}  
    \definecolor{TUMWhite}           {cmyk}{0.00,0.00,0.00,0.00}  
    \definecolor{TUMBlue}            {cmyk}{1.00,0.43,0.00,0.00}  
    \definecolor{TUMDarkBlue}        {cmyk}{1.00,0.57,0.12,0.70}  
    \definecolor{TUMDarkerBlue}      {cmyk}{1.00,0.54,0.04,0.19}  
    \definecolor{TUMMediumBlue}      {cmyk}{0.90,0.48,0.00,0.00}  
    \definecolor{TUMLighterBlue}     {cmyk}{0.65,0.19,0.01,0.04}  
    \definecolor{TUMLightBlue}       {cmyk}{0.42,0.09,0.00,0.00}  
    \definecolor{TUMDarkGray}        {cmyk}{0.00,0.00,0.00,0.80}  
    \definecolor{TUMMediumGray}      {cmyk}{0.00,0.00,0.00,0.50}  
    \definecolor{TUMLightGray}       {cmyk}{0.00,0.00,0.00,0.20}  
    \definecolor{TUMGreen}           {cmyk}{0.35,0.00,1.00,0.20}  
    \definecolor{TUMOrange}          {cmyk}{0.00,0.65,0.95,0.00}  
    \definecolor{TUMIvory}           {cmyk}{0.03,0.04,0.14,0.08}  
    \definecolor{TUMBlack}           {rgb} {0.000,0.000,0.000}    
    \definecolor{TUMWhite}           {rgb} {1.000,1.000,1.000}    
    \definecolor{TUMBlue}            {rgb} {0.000,0.396,0.741}    
    \definecolor{TUMDarkBlue}        {rgb} {0.000,0.200,0.349}    
    \definecolor{TUMDarkerBlue}      {rgb} {0.000,0.322,0.576}    
    \definecolor{TUMMediumBlue}      {rgb} {0.000,0.451,0.812}    
    \definecolor{TUMLighterBlue}     {rgb} {0.392,0.627,0.784}    
    \definecolor{TUMLightBlue}       {rgb} {0.596,0.776,0.918}    
    \definecolor{TUMDarkGray}        {rgb} {0.345,0.345,0.353}    
    \definecolor{TUMMediumGray}      {rgb} {0.612,0.616,0.624}    
    \definecolor{TUMLightGray}       {rgb} {0.851,0.855,0.859}    
    \definecolor{TUMGreen}           {rgb} {0.635,0.678,0.000}    
    \definecolor{TUMOrange}          {rgb} {0.890,0.447,0.133}    
    \definecolor{TUMIvory}           {rgb} {0.855,0.843,0.796}    
    \definecolor{TUMBeamerYellow}    {rgb} {1.000,0.706,0.000}    
    \definecolor{TUMBeamerOrange}    {rgb} {1.000,0.502,0.000}    
    \definecolor{TUMBeamerRed}       {rgb} {0.898,0.204,0.094}    
    \definecolor{TUMBeamerDarkRed}   {rgb} {0.792,0.129,0.247}    
    \definecolor{TUMBeamerBlue}      {rgb} {0.000,0.600,1.000}    
    \definecolor{TUMBeamerLightBlue} {rgb} {0.255,0.745,1.000}    
    \definecolor{TUMBeamerGreen}     {rgb} {0.569,0.675,0.420}    
    \definecolor{TUMBeamerLightGreen}{rgb} {0.710,0.792,0.510}    

\usepackage{helvet}
\usepackage[eulergreek]{sansmath}

\newcommand{\cb}[1]{\textcolor{black}{#1}}

\newcommand{\new}[1]{\textcolor{black}{#1}}

\copyrightnotice{U.S.\ Government work not protected by U.S.\ copyright}

\copyrightnotice{979-8-3503-2411-2/23/\$31.00 {\copyright}2023 Crown}

\copyrightnotice{979-8-3503-2411-2/23/\$31.00 {\copyright}2023 European Union}

\copyrightnotice{979-8-3503-2411-2/23/\$31.00 {\copyright}2023 IEEE}

\toappear{Preprint submitted to 2023 IEEE International Workshop on Machine Learning for Signal Processing}

\title{Reverse Ordering Techniques for Attention-based Channel Prediction} 

\name{Valentina Rizzello, Benedikt B\"{o}ck, Michael Joham, Wolfgang Utschick}
\address{Department of Computer Engineering, Technical University of Munich}

\usetikzlibrary{shapes.geometric}
\begin{document}

\ninept

\maketitle

\begin{abstract}
This work aims to predict channels in wireless communication systems based on noisy observations, utilizing sequence-to-sequence models with attention (Seq2Seq-attn) and transformer models. Both models are adapted from natural language processing to tackle the complex challenge of channel prediction. Additionally, a new technique called reverse positional encoding is introduced in the transformer model to improve the robustness of the model against varying sequence lengths. Similarly, the encoder outputs of the Seq2Seq-attn model are reversed before applying attention. Simulation results demonstrate that the proposed ordering techniques allow the models to better capture the relationships between the channel snapshots within the sequence, irrespective of the sequence length, as opposed to existing methods.
\end{abstract}
\section{Introduction}
\label{sec:intro}
In 5G, and beyond 5G, wireless communication systems, the \ac{csi} is essential for the \ac{bs} to optimize its transmission strategy to communicate to the receiving \ac{mt}. 
The \ac{csi}, or channel, is a complex-valued matrix whose dimensions correspond to the number of transmit and receive antennas. It describes the link between each transmit and receive antenna pair, that can be affected by factors such as fading, multipath propagation, and interference from other signals. In a typical \ac{fdd} system, the \ac{bs} sends a predefined sequence of symbols, called pilots, to the \ac{mt}, which estimates the \ac{csi} and feeds the \ac{csi} coefficients back to the \ac{bs}. Hence, there is an inevitable delay between the instant of when the \ac{mt} estimates the \ac{csi} and the one in which the \ac{bs} receives the \ac{csi} coefficient. Therefore, since the channels change over time, it is crucial for the \ac{bs} to predict the channel.
The problem of channel prediction is quite straightforward when the channel dynamics are known.
In particular, when the Doppler frequency is known, linear predictors such as \ac{ar} models or \acp{kf} can effectively be used for tracking the \ac{csi}, see~\cite{TWC_BaddourB05,TWC_BarbieriPC09,DSP_HusseiniSR17,zemen-prediction}.
However, in a typical wireless communication system, the \acp{mt} move with unknown channel statistics and different velocities. Therefore, a finite number of \ac{ar} predictors need to be pre-trained for different Doppler frequencies and the channel parameters must be estimated from the available data. In order for this approach to work well, both, \textit{i)} the Doppler frequency must be correctly estimated, and \textit{ii)} a potentially large number of linear predictors need to be stored. Additionally, a wrong or a coarse approximation of the Doppler frequency can cause a non-negligible loss.

In recent years, \acp{nn} have become a promising solution in various research fields including wireless communications. In~\cite{Yuan2020}, \acp{cnn} are used in combination with \ac{ar} models for \ac{csi} forecasting. In particular, \acp{cnn} are used to correctly identify the channel dynamics, and to load the corresponding pre-trained \ac{ar} predictor to forecast the CSI. The authors of~\cite{Pratik2021} also propose a hybrid approach called Hypernetwork Kalman Filter. There, a single-antenna setup is considered and only Kalman equations are utilized for prediction, whereas a hypernetwork continuously updates the Kalman parameters based on past observations.
In~\cite{Yuan2020,JiangSS20,Shehzad2022} \acp{rnn} are used for \ac{csi} prediction. In particular, due to the ability of \acp{rnn} to incorporate the  typical dynamics of time series data, they represent a valid alternative to \ac{ar} models for time series forecasting. However, notably \acp{rnn} are difficult to train due to vanishing or exploding gradient issues, see~\cite{PascanuMB13}.
Among the recent advances, we find the work in~\cite{Chu2022} where the \ac{csi} prediction is incorporated in a reinforcement learning-based setup with goal to maximize the multi-user sum rate over time. In this setting, the so-called Actor Network is responsible for \ac{csi} prediction and it is realized via a \ac{mlp}. 
The most recent study in~\cite{transformer-parallel} presents a novel approach to predict future channels in parallel using a transformer-based parallel channel prediction scheme. The main objective is to prevent error propagation that often occurs in sequential prediction. To this end, the one-step ahead sequential prediction in the transformer decoder, also called dynamic decoding in the literature, is eliminated completely. Instead, the transformer decoder takes as input a certain number of the past channel realizations, along with a specific number of all-zero vectors equal to the number of unknown channels, to predict all the future channels in parallel.

In this study, we draw inspiration from the achievements in natural language processing by so-called attention-based models. \cb{We adapt both, the transformer and the \ac{seqa} models, to the channel prediction task. Instead of settling for the vanilla architecture~\cite{VaswaniSPUJGKP17,Wu2020}, we introduce a novel \ac{rpe} technique in the transformer model to improve the model's robustness against variable sequence lengths during testing, that may differ from the lengths assumed for training. With the same goal in mind, we also reverse the encoder outputs of the \ac{seqa} model before applying attention.
Unlike the state of the art, we investigate the challenging setup where only noisy channels are available for training and where the users are moving inside a cell within a wide range of velocities, i.e., between 0~km/h and 120~km/h. We evaluate our models for varying noise levels and sequence lengths, including lengths that differ from those used during training.
Simulation results show that the proposed models exhibit solid performance for different sequence lengths.} 
\new{Our main contributions are \textit{i)} adapting both the transformer and the \ac{seqa} model to the channel prediction task; \textit{ii)} introducing novel ordering techniques in those models to make them robust in adapting to \ac{csi} sequences of any length, and therefore reducing complexity and storage requirements at the \ac{bs}.}

The rest of the paper is organized as follows. In Section~\ref{sec:sys-model}, the system model is described; in Section~\ref{sec:tnn}, the proposed Transformer-RPE model is presented; in Section~\ref{sec:seq}, the proposed \ac{seqa}-R model is presented; in Section~\ref{sec:sims}, the dataset used and the training setup are presented, and the simulation results are discussed; in Section~\ref{sec:concl}, we draw our conclusions.

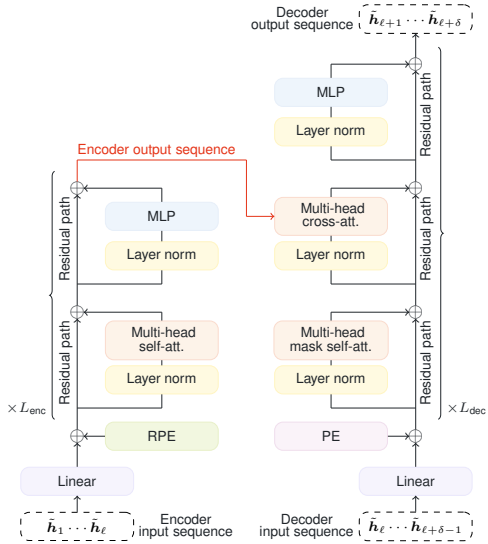
\begin{figure}[t]
    \centering
    \scalebox{0.75}{\scalebox{1.0}{\begin{tikzpicture}[tum-grey-1]

\node [input, name=input] at (0, 0) {};
\node[draw=black, dashed, minimum height=0.5cm, minimum width=2cm, rectangle, rounded corners=0.15cm](ch1) at (0, -0.8){\scalebox{1}{\textcolor{black}{\scriptsize{$\tilde{\vh}_1 \cdots \tilde{\vh}_{\ell}$}}}};
\node[right=-0.2cm of ch1]{$\def\arraystretch{0.4}\begin{array}{c} \text{\sansmath\sffamily\scriptsize{Encoder}} \\ \text{\sansmath\sffamily\scriptsize{input sequence}}\end{array} $ };
\node[draw=tum-blue-bright-3!60!white, fill=tum-blue-bright-4!60!white, minimum height=0.5cm, minimum width=2cm, rectangle, rounded corners=0.15cm](lin1) at (0, 0){\sffamily\scriptsize{Linear}};
\node [circle, fill=white, inner sep=-1.4pt] (plus1) at (0, 0.8) {$\oplus$};
\node [draw=tum-green-3!60!white, fill=tum-green-4!60!white,minimum height=0.5cm, minimum width=2cm, rectangle, rounded corners=0.15cm, ](rev-pos) at (1.5, 0.8) {\sffamily\scriptsize{RPE}};
\draw[->, >=to](lin1)--(plus1);
\draw[->, >=to](rev-pos)--(plus1);
\node[draw=tum-yellow-3!60!white, fill=tum-yellow-4!60!white, minimum height=0.5cm, minimum width=2cm, rectangle, rounded corners=0.15cm](ln1) at (1.5, 1.8) {\sffamily\scriptsize{Layer norm}};
\node[draw=tum-red-3!60!white, fill=tum-red-4!60!white, minimum height=0.5cm, minimum width=2cm, rectangle, rounded corners=0.15cm](mh1) at (1.5, 2.5) {$\def\arraystretch{0.5}\begin{array}{c} \text{\sansmath\sffamily\scriptsize{ Multi-head}} \\\text{\sansmath\sffamily\scriptsize{self-att.}}\end{array} $ };
\node[draw=tum-yellow-3!60!white, fill=tum-yellow-4!60!white, minimum height=0.5cm, minimum width=2cm, rectangle, rounded corners=0.15cm](ln2) at (1.5, 4.0) {\sffamily\scriptsize{Layer norm}};
\node[draw=tum-blue-light-3!60!white, fill=tum-blue-light-4!60!white, minimum height=0.5cm, minimum width=2cm, rectangle, rounded corners=0.15cm](mlp1) at (1.5, 4.7) {\sffamily\scriptsize{MLP}};
\node [circle, fill=white, inner sep=-1.4pt] (plus2) at (0, 3) {$\oplus$};
\node [circle, fill=white, inner sep=-1.4pt] (plus3) at (0, 5.2) {$\oplus$};
\draw[->, >=to](plus1)--(plus2);
\draw[->, >=to](plus2)--(plus3);

\coordinate (dm1) at (1.5, 1.3);
\coordinate (dm0) at (0, 1.3);
\draw (dm1)--(ln1);
\draw (dm0)--(dm1);
\draw (ln1)--(mh1);
\coordinate (dm2) at (1.5, 3);
\draw (mh1)--(dm2);
\draw[->, >=to](dm2)--(plus2);

\coordinate (dm4) at (1.5, 3.5);
\coordinate (dm3) at (0, 3.5);
\coordinate (dm5) at (1.5, 5.2);
\draw[->, >=to](dm5)--(plus3);
\draw (dm3)--(dm4);
\draw (dm4)--(ln2);
\draw (ln2)--(mlp1);
\draw (mlp1)--(dm5);

\coordinate (dm6) at (0, 6.2);

\draw[->, >=to](ch1)--(lin1);

\node[draw=black, dashed, minimum height=0.5cm, minimum width=2cm, rectangle, rounded corners=0.15cm](ch2) at (6, -0.8){\scalebox{1}{\textcolor{black}{\scriptsize{$\tilde{\vh}_{\ell} \cdots \tilde{\vh}_{\ell + \delta - 1}$}}}};
\node[left=-0.2cm of ch2]{$\def\arraystretch{0.4}\begin{array}{c} \text{\sansmath\sffamily\scriptsize{Decoder}} \\ \text{\sansmath\sffamily\scriptsize{input sequence}}\end{array} $ };

\node[draw=tum-blue-bright-3!60!white, fill=tum-blue-bright-4!60!white, minimum height=0.5cm, minimum width=2cm, rectangle, rounded corners=0.15cm](dlin1) at (6, 0) {\sffamily\scriptsize{Linear}};
\node[draw=tum-yellow-3!60!white, fill=tum-yellow-4!60!white,minimum height=0.5cm, minimum width=2cm, rectangle, rounded corners=0.15cm](dln1) at (4.5, 1.8) {\sffamily\scriptsize{Layer norm}};
\node[draw=tum-red-3!60!white, fill=tum-red-4!60!white, minimum height=0.5cm, minimum width=2cm, rectangle, rounded corners=0.15cm](dmh1) at (4.5, 2.5) {$\def\arraystretch{0.5}\begin{array}{c} \text{\sansmath\sffamily\scriptsize{ Multi-head}} \\\text{\sansmath\sffamily\scriptsize{mask~self-att.}}\end{array} $ };
\node [draw=tum-pink-3!60!white, fill=tum-pink-4!60!white, minimum height=0.5cm, minimum width=2cm, rectangle, rounded corners=0.15cm](pos) at (4.5, 0.8) {\sffamily\scriptsize{PE}};
\node [circle, fill=white, inner sep=-1.4pt] (dplus1) at (6, 0.8) {$\oplus$};

\node[draw=tum-yellow-3!60!white, fill=tum-yellow-4!60!white, minimum height=0.5cm, minimum width=2cm, rectangle, rounded corners=0.15cm](dec-ln2) at (4.5, 4.0) {\sffamily\scriptsize{Layer norm}};
\node[draw=tum-red-3!60!white, fill=tum-red-4!60!white, minimum height=0.5cm, minimum width=2cm, rectangle, rounded corners=0.15cm](enc-dec-att) at (4.5, 4.7){$\def\arraystretch{0.5}\begin{array}{c} \text{\sansmath\sffamily\scriptsize{ Multi-head}} \\\text{\sansmath\sffamily\scriptsize{cross-att.}}\end{array} $ };
\node [circle, fill=white, inner sep=-1.4pt] (dplus2) at (6, 3) {$\oplus$};

\node[draw=tum-yellow-3!60!white, fill=tum-yellow-4!60!white,minimum height=0.5cm, minimum width=2cm, rectangle, rounded corners=0.15cm](dec-ln3) at (4.5, 6.2) {\sffamily\scriptsize{Layer norm}};
\node[draw=tum-blue-light-3!60!white, fill=tum-blue-light-4!60!white, minimum height=0.5cm, minimum width=2cm, rectangle, rounded corners=0.15cm](dec-mlp) at (4.5, 6.9) {\sffamily\scriptsize{MLP}};

\node [circle, fill=white, inner sep=-1.4pt] (dplus3) at (6, 5.2) {$\oplus$};
\node [circle, fill=white, inner sep=-1.4pt] (dplus4) at (6, 7.4) {$\oplus$};

\draw[->, >=to] (pos)--(dplus1);
\draw[->, >=to] (dlin1)--(dplus1);
\draw[->, >=to] (ch2)--(dlin1);
\draw[->, >=to] (dplus1)--(dplus2);
\draw[->, >=to] (dplus2)--(dplus3);
\draw[->, >=to] (dplus3)--(dplus4);

\coordinate (p0) at (6, 1.3);
\coordinate (p1) at (4.5, 1.3);
\coordinate (p2) at (4.5, 3);
\draw (p0)--(p1);
\draw[->, >=to] (p2)--(dplus2);
\draw (p1)--(dln1);
\draw (dln1)--(dmh1);
\draw (dmh1)--(p2);

\coordinate (q0) at (4.5, 3.5);
\coordinate (q1) at (6, 3.5);
\coordinate (q2) at (4.5, 5.2);
\draw (q0)--(q1);
\draw[->, >=to] (q2)--(dplus3);
\draw (q0)--(dec-ln2);
\draw (dec-ln2)--(enc-dec-att);
\draw (enc-dec-att)--(q2);

\coordinate (w0) at (4.5, 5.7);
\coordinate (w2) at (4.5, 7.4);
\coordinate (w1) at (6, 5.7);
\draw (w0)--(w1);
\draw[->, >=to] (w2)--(dplus4);
\draw (w0)--(dec-ln3);
\draw (dec-ln3)--(dec-mlp);
\draw (dec-mlp)--(w2);

\node[draw=black, dashed, minimum height=0.5cm, minimum width=2cm, rectangle, rounded corners=0.15cm](ch3) at (6, 8.2){\scriptsize{\textcolor{black}{$\tilde{\vh}_{\ell + 1} \cdots \tilde{\vh}_{\ell + \delta}$}}};
\node[left=-0.2cm of ch3]{$\def\arraystretch{0.4}\begin{array}{c} \text{\sansmath\sffamily\scriptsize{Decoder}} \\ \text{\sansmath\sffamily\scriptsize{output sequence}}\end{array} $ };

\draw [->, >=to] (dplus4)--(ch3);

\coordinate (i0) at (0, 5.7);
\coordinate (i1) at (3, 5.7);
\coordinate (i2) at (3, 4.7);

\draw[color=TUMBeamerRed] (i0)--(i1);
\draw[color=TUMBeamerRed] (i1)--(i2);
\draw[->, >=to, color=TUMBeamerRed] (i2)--(enc-dec-att);
\draw[color=TUMBeamerRed] (plus3)--(i0);

\node[above left=-0.05cm and -0.1cm of i1]{\textcolor{TUMBeamerRed}{$\def\arraystretch{0.4}\begin{array}{c} \text{\sansmath\sffamily\scriptsize{Encoder output sequence}}\end{array} $} };

\node[] (text1) at (-0.2, 2.1){\rotatebox{90}{\sansmath\sffamily\scriptsize{Residual path}}};
\node[] (text2) at (-0.2, 4.3){\rotatebox{90}{\sansmath\sffamily\scriptsize{Residual path}}};
\node[] (text3) at (6.2, 2.1){\rotatebox{90}{\sansmath\sffamily\scriptsize{Residual path}}};
\node[] (text4) at (6.2, 4.3){\rotatebox{90}{\sansmath\sffamily\scriptsize{Residual path}}};
\node[] (text5) at (6.2, 6.5){\rotatebox{90}{\sansmath\sffamily\scriptsize{Residual path}}};

\draw [decorate, decoration = {brace,}] (-0.4, 1.1) --  (-0.4, 5.5);
\draw [decorate, decoration = {brace, mirror}] (6.4, 1.1) --  (6.4, 7.7);
\node[] at (-0.9, 1.3) {\scalebox{1}{\scriptsize{$\times L_{\text{\sffamily{enc}}}$}}};
\node[] at (6.9, 1.3) {\scalebox{1}{\scriptsize{$\times L_{\text{\sffamily{dec}}}$}}};

\end{tikzpicture}}}
    \caption{Transformer-RPE for \ac{csi} prediction.}
  \label{fig:tnn_fig}
\end{figure}

\section{System model}
\label{sec:sys-model}
We consider a \ac{bs} serving multiple \acp{mt} in a typical 5G cell. The \ac{bs} is equipped with $M$ antennas, whereas the single-antenna users are moving with different velocities and, therefore, experience different fading conditions.
In particular, we assume that the channel remains constant for the duration of a slot which we denote as
$T_{\text{slot}}$, and that a frame contains $N_{\text{slot}}$ slots.
Additionally, we assume that the velocities of the users are constant within the duration of a frame.
From now on, we denote as $\vh_i \in \mathbb{C}^M$ the \ac{csi} vector corresponding to the $i$th generic slot of the CSI time series. The sequence of $N_{\text{slot}}$ subsequent \ac{csi} vectors $\{\vh_i\}_{i=1}^{N_{\text{slot}}}$ is assumed to be strongly correlated. The goal of multi-step \ac{csi} prediction is to find the best estimator $f: \mathbb{C}^{M \times \ell} \rightarrow \mathbb{C}^{M \times \delta}$ which predicts the \ac{csi} vectors $\{\vh_i\}_{i=\ell + 1}^{\ell + \delta}$ based on preceding $l$ observations $\{\vh_i\}_{i=1}^{\ell}$ with $\ell + \delta \leq N_{\text{slot}}$. 
Note that, we assume that the channels $\{\vh_i\}_{i=1}^{\ell}$ are not perfectly known, and that only the corresponding noisy observations $\{\check{\vh}_i\}_{i=1}^{\ell}$ are available, i.e.,
\begin{equation}
\label{eq:noisy-obs}
    \check{\vh}_i \gets \vh_i + \vn_i \quad i=1, \dots \ell,
\end{equation}
where $\vn_i$ denotes the complex-valued noise vector with independent elements distributed as $\mathcal{N}_{\mathbb{C}}(0, \sigma_n^2)$, and such that $\mathbb{E}[\vn_i \vn_j^{\He}]= \boldsymbol{0} $ for all $i \neq j$.
In addition, throughout this study, we consider real-valued neural networks. Therefore, we transform the complex vector $\vh_i$ into a real vector where the real and imaginary parts of the original vector are concatenated as
\begin{equation}
\label{eq: toreal}
    \mathbb{R}^{2M} \ni \tilde{\vh}_i = \mathrm{concat}(\mathfrak{Re}(\check{\vh}_i), \mathfrak{Im}(\check{\vh}_i)). 
\end{equation}

\section{Transformer-RPE model}
\label{sec:tnn}
\begin{algorithm}[t]
\footnotesize{\caption{Multi-head (masked) self- or cross- attention~\cite{VaswaniSPUJGKP17}}\label{alg:mha}
\begin{algorithmic}
\Require $\vX \in \mathbb{R}^{d_{\text{x}} \times l_{\text{x}}}$, $\vZ \in \mathbb{R}^{d_{\text{z}} \times l_{\text{z}}}$, $\mathrm{Mask} \in \{0,1\}^{l_{\text{z}} \times l_{\text{x}}}$, primary sequence, context sequence, and an optional mask
\Ensure $\vY \in \mathbb{R}^{d_{\text{out}} \times l_{\text{x}}} $, updated representation of $\vX$
\renewcommand{\algorithmicensure}{\textbf{Hyperparameters:}}
\Ensure $H$, number of attention-heads 
\renewcommand{\algorithmicensure}{\textbf{Learnable parameters:}}
\Ensure $\vW_{\text{q}} \in \mathbb{R}^{H d_{\text{attn}} \times d_{\text{x}}}$,  $\vW_{\text{k}} \in \mathbb{R}^{H d_{\text{attn}} \times d_{\text{z}}}$, $\vW_{\text{v}} \in \mathbb{R}^{H d_{\text{mid}} \times d_{\text{z}}}$, $\vW_{\text{o}} \in \mathbb{R}^{d_{\text{out}} \times H d_{\text{mid}}}$
\renewcommand{\algorithmicensure}{\textbf{Output:}}
\State $\vQ \gets \vW_{\text{q}}\vX$ \Comment{queries $\in \mathbb{R}^{H d_{\text{attn}} \times l_{\text{x}}}$}

\State $\vK \gets \vW_{\text{k}}\vZ$ \Comment{keys $\in \mathbb{R}^{H d_{\text{attn}} \times l_{\text{z}}}$}

\State $\vV \gets \vW_{\text{v}}\vZ$ \Comment{values $\in \mathbb{R}^{H d_{\text{mid}} \times l_{\text{z}}}$}

\For{$h=1$ to $H$}
\State $\vS^{(h)} \gets \vK^{(h),\Tr}\vQ^{(h)}$ \Comment{scores $\in \mathbb{R}^{l_{\text{z}} \times l_{\text{x}}}$}
\If{$\mathrm{Mask}$} \State $\vS^{(h)}[\neg \mathrm{Mask}] \gets -\infty$ \EndIf
\State $\tilde{\vV}^{(h)} \gets \vV^{(h)} \cdot $\texttt{softmax}$(\vS^{(h)} / \sqrt{d_{\text{attn}}})$ \Comment{$\in \mathbb{R}^{ d_{\text{mid}} \times l_{\text{x}}}$}
\EndFor
\State $\tilde{\vV} \gets [ \tilde{\vV}^{(1)}; \dots; \tilde{\vV}^{(H)}] $\Comment{$\in \mathbb{R}^{H d_{\text{mid}} \times l_{\text{x}}}$}
\State $\vY \gets \vW_{\text{o}}\tilde{\vV}$ \Comment{$\in \mathbb{R}^{ d_{\text{out}} \times l_{\text{x}}}$}
\end{algorithmic}}
\end{algorithm}

In this section, we describe the proposed Transformer with \ac{rpe} model for CSI prediction, which has the Transformer model~\cite{VaswaniSPUJGKP17} as baseline. The Transformer-\ac{rpe} model is illustrated in Fig.~\ref{fig:tnn_fig}. In the following, we provide a brief description of the Transformer model. However, the reader can refer to~\cite{VaswaniSPUJGKP17} for a more detailed explanation.
A Transformer consists of an \textit{encoder} and a \textit{decoder} \ac{nn}. The encoder aims to extract the important information from its input sequence, which can help the decoder to predict the next slots one by one in a subsequent step. This input sequence is represented by the $\ell$ known noisy channels $\{\tilde{\vh}_i\}_{i=1}^{\ell}$.
The encoder has $L_{\text{enc}}$ layers and each layer contains two consecutive residual networks. The first residual network has the ``multi-head attention'' as the main layer, whereas the second residual network contains an \ac{mlp} as the main module. Moreover, the \ac{ln}~\cite{layer-norm} precedes each of these modules. The \ac{mlp} contains two fully-connected layers and a GeLU~\cite{gelu} activation function after the first layer.
A concise pseudo-code of the multi-head attention layer can be found in Algorithm~\ref{alg:mha}. In the transformer encoder, we compute a multi-head self-attention, which means that in Algorithm~\ref{alg:mha}, the context sequence $\vZ$ is equal to the primary sequence $\vX$.\\
The decoder has $L_{\text{dec}}$ layers and each layer contains three consecutive residual networks. The first and the second residual networks have the multi-head attention layer as the main module of the residual block, whereas the third residual network contains an \ac{mlp} as the main module. As for the encoder layer, the \ac{ln} precedes each of these modules. 
In the first residual network of the decoder, Algorithm~\ref{alg:mha} takes as input, in addition to the primary sequence $\vX$, a mask that ensures that the prediction of $\vh_{\ell+k}$ only depends on $\{\vh_{\ell+j}\}_{j=0}^{k-1}$. In the Algorithm~\ref{alg:mha}, this is achieved by setting all the values of the input of the $\softmax$ activation function~\cite{eslii} to $- \infty$ which correspond to ``illegal'' connections. The primary and context sequence coincide. On the contrary, in the second residual network of the decoder, Algorithm~\ref{alg:mha} takes the encoder output sequence as context sequence $\vZ$. In this way, the information contained in the known \ac{csi} slots can be leveraged to predict the next slot.

For the proposed model, the first element of the decoder input sequence is the last known CSI snapshot $\tilde{\vh}_{\ell}$. This is in contrast to what happens in natural language processing since there, due to the absence of previous output at the beginning of a translation, a predefined start-of-sentence token is given as first decoder input.
Additionally, during training, teacher-forcing~\cite{WilliamsZ89} is used in the decoder, which means that during training, we use the true noisy \ac{csi} observation $\tilde{\vh}_{\ell+1}, \dots, \tilde{\vh}_{\ell + \delta -1}$ as further decoder inputs, instead of the predicted ones obtained at the decoder output.
This helps to speed up the training process since the decoder outputs $\tilde{\vh}_{\ell+1}, \dots, \tilde{\vh}_{\ell + \delta}$ can be obtained in parallel. However, during testing, we have to stick to a sequential one-by-one prediction.

Similarly to the original implementation in~\cite{VaswaniSPUJGKP17} also in this case, we transform the input sequences of both, the encoder and the decoder, first by a linear layer and then by adding a constant bias term, called \ac{pe}. The \ac{pe} consists of constant, non-learnable vectors that are added after the first linear layer in both, the encoder and the decoder. Since there is no recurrence in the model, the \ac{pe} is the only way to inject information about the order of the sequence. Therefore, it is crucial for the transformer architecture. In~\cite{VaswaniSPUJGKP17}, and in our work, the \ac{pe} is constructed with sine and cosine functions as
\begin{equation}
\label{eq: pos-emb}
\begin{gathered}
        \textrm{PE}(j, 2i) = \sin(j/(10000^{2i/d_{\text{model}}}))\\
        \textrm{PE}(j, 2i + 1) = \cos(j/(10000^{2i/d_{\text{model}}})),
\end{gathered}
\end{equation}
where $j \in \{0, \ell-1\}$ is the index corresponding to the position within the sequence, $i \in \{0, \lfloor d_{\text{model}}/2 \rfloor \}$ is the index of the dimension, and $d_{\text{model}}$ corresponds to the dimension of each \ac{csi} slot after the first linear layer.
\cb{However, differently from the vanilla architecture, we introduce a novel \ac{rpe} in the transformer encoder, while keeping the standard \ac{pe} in the transformer decoder. Intuitively, this means that we start counting the \ac{csi} snapshots in the encoder from the last known snapshot. In other words, this enhances the robustness of the transformer to sequences of variable lengths, as the \ac{pe} linked to the latest known slots remains consistent for shorter or longer sequences. The \ac{rpe} can be obtained by first computing the standard \ac{pe} in~\eqref{eq: pos-emb} and then by reversing the order with respect to the position index $j$. To better understand the motivation behind this procedure, we can consider a simple example in which the transformer~\cite{VaswaniSPUJGKP17} with standard \ac{pe} in the encoder is trained with an encoder input sequence of length $\ell$, whereas it is tested with an encoder input sequence of length $\nu$, with $\nu \neq \ell$. 
During training, the transformer implicitly exploits the information contained in the most recent snapshots more than the information contained in the initial snapshots to make a good prediction. In the transformer with standard \ac{pe} those are the snapshots associated with, e.g., $\textrm{PE}(\ell - 1, :), \dots, \textrm{PE}(\ell - \zeta, :)$ with $\zeta < \ell$, and the colon that denotes all the elements of the corresponding row. However, when $\nu < \ell$, such model fails to make a good prediction because in this case the most important (or recent) snapshots are associated with the \acp{pe} that were linked with the initial snapshots during training, thus their importance is underestimated. On the other hand, when $\nu > \ell$, the usual ordering leads to the situation in which those snapshots, which are now outdated, are interpreted as the most recent ones by the model, and their importance is overestimated for making prediction.
This problem is solved with the proposed \ac{rpe}, which introduces consistency when mapping the \acp{pe} to the corresponding snapshots, and makes the transformer robust, allowing to capture the relationships between different snapshots in the sequence, regardless of the sequence length.}

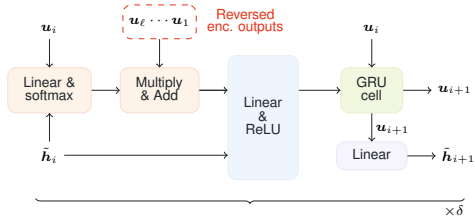
\begin{figure}[t]
\centering
  \scalebox{0.75}{\scalebox{1}{\begin{tikzpicture}[tum-grey-1]
\node[input, name=input]{};

\node[draw=tum-red-3!60!white, fill=tum-red-4!60!white, minimum height=0.8cm, minimum width=0.8cm, rectangle, rounded corners=0.15cm](cell1) {{$\def\arraystretch{0.5}\begin{array}{c} \text{\sansmath\sffamily\scriptsize{Linear \&}} \\ \text{\sansmath\sffamily\scriptsize{softmax}} \end{array} $}};

\node[above=0.5cm of cell1, name=n1]{\textcolor{black}{\scriptsize{$\vu_{i}$}}};
\draw[->, >=to](n1)--(cell1);

\node[right=0.5cm of cell1, name=n2]{};
\draw[->, >=to](cell1)--(n2);

\node[below=0.5cm of cell1, name=n3]{\textcolor{black}{\scriptsize{$\tilde{\vh}_{i}$}}};
\draw[->, >=to](n3)--(cell1);

\node[draw=tum-red-3!60!white, fill=tum-red-4!60!white, minimum height=0.8cm, minimum width=0.8cm, rectangle, rounded corners=0.15cm, right=0.5cm of cell1](cell2) {{$\def\arraystretch{0.5}\begin{array}{c} \text{\sansmath\sffamily\scriptsize{Multiply}} \\ \text{\sansmath\sffamily\scriptsize{\& Add}} \end{array} $}};

\node[right=0.5cm of cell2, name=n4]{};
\draw[->, >=to](cell2)--(n4);

\node[above=0.5cm of cell2, draw=TUMBeamerRed, dashed, minimum height=0.6cm, minimum width=1.2cm, rectangle, rounded corners=0.15cm, name=n5]{\textcolor{black}{\scriptsize{$\vu_{\ell} \cdots \vu_1$}}};
\node[right=-0.1cm of n5]{\textcolor{TUMBeamerRed}{{$\def\arraystretch{0.5}\begin{array}{c} \text{\sansmath\sffamily\scriptsize{Reversed}} \\ \text{\sansmath\sffamily\scriptsize{enc. outputs}} \end{array} $}}};
\draw[->, >=to](n5)--(cell2);

\node[draw=tum-green-3!60!white, fill=tum-green-4!60!white, minimum height=0.8cm, minimum width=0.8cm, rectangle, rounded corners=0.15cm, right=2.5cm of cell2](cell4) {{$\def\arraystretch{0.5}\begin{array}{c} \text{\sansmath\sffamily\scriptsize{GRU}}\\ \text{\sansmath\sffamily\scriptsize{cell}} \end{array} $}};

\draw[->, >=to](cell2)--(cell4);

\node[draw=tum-blue-light-3!60!white, fill=tum-blue-light-4!60!white, minimum height=2.2cm, minimum width=0.5cm, rectangle, rounded corners=0.15cm, below right=-1.03cm and 0.5cm of cell2](cell3) {{$\def\arraystretch{0.5}\begin{array}{c} \text{\sansmath\sffamily\scriptsize{Linear}} \\ \text{\sansmath\sffamily\scriptsize{\&}} \\ \text{\sansmath\sffamily\scriptsize{ReLU}} \end{array} $}};

\node[right=4cm of n3, name=conn2]{};
\node[below=1cm of n4, name=conn1]{};
\node[name=inter1] at (intersection of  n3--conn2 and n4--conn1){};

\draw[->, >=to](n3)--(inter1);

\node[right=0.5cm of cell4, name=n9]{\textcolor{black}{\scriptsize{${\vu}_{i + 1}$}}};
\draw[->, >=to](cell4)--(n9);

\node[above=0.5cm of cell4, name=n6]{\textcolor{black}{\scriptsize{$\vu_{i}$}}};
\draw[->, >=to](n6)--(cell4);

\node[draw=tum-blue-bright-3!60!white, fill=tum-blue-bright-4!60!white, minimum height=0.5cm, minimum width=0.8cm, rectangle, rounded corners=0.15cm, below=0.5cm of cell4](cell5) {{$\def\arraystretch{0.5}\begin{array}{c} \text{\sansmath\sffamily\scriptsize{Linear}} \end{array} $}};

\draw[->, >=to](cell4)--(cell5);

\node[below right=0.1cm and -0.6cm of cell4]{\textcolor{black}{\scriptsize{${\vu}_{i + 1}$}}};

\node[right=0.5cm of cell5, name=n7]{\textcolor{black}{\scriptsize{$\tilde{\vh}_{i + 1}$}}};
\draw[->, >=to](cell5)--(n7);

\node[below left=1.1cm and -0.5cm of cell1, name=s1]{};
\node[right=7.6cm of s1, name=s2]{};
\draw [
    decoration={
        brace,
        mirror,
        raise=0.3cm
    },
    decorate
] (s1) -- (s2); 
\node[below right=0.2cm and -0.7cm of s2]{\textcolor{black}{\scriptsize{$\times \delta$}}};

\end{tikzpicture}}}
  \caption{Decoder of Seq2Seq-attn-R model for CSI prediction.}
  \label{fig:seq2seq} 
\end{figure}

\section{Seq2Seq-Attn-R model}
\label{sec:seq}
Another relevant framework to solve the problem at hand is the \ac{seq} architecture, see~\cite{SutskeverVL14}.
Like the transformer, also the \ac{seq} model comprises an \textit{encoder} and a \textit{decoder} neural network, which are both \acp{rnn} in the simplest case. Specifically, the encoder \ac{rnn} encodes the input sequence to produce a final state which in turn is used as initial state for the decoder \ac{rnn}. The hope is that the final state of the encoder encodes all the important information about the source or input sequence such that the decoder can generate the target sequence based on this vector. 
However, in such setting, the decoder has to extract meaningful information from a single representation (the final state of the encoder), which can be a daunting task, especially when taking into account long sequences, or sentences. In~\cite{BahdanauCB14} an attention mechanism has been introduced in the decoder neural network to address this problem. In particular, instead of passing only the final state of the encoder \ac{rnn}, this approach involves passing all the encoder \ac{rnn} states to the decoder. Hence, at each decoder step, the attention mechanism decides which parts of the source sequence are more relevant.

In the following, we propose an adapted model of~\cite{BahdanauCB14}, called \ac{seqa}-R model, to tackle the channel prediction task.
To avoid vanishing or exploding gradient problems,\footnote{Because \acp{rnn} allow for information to be fed back to the same node multiple times, they are prone to vanishing and exploding gradient problems. The feedback can cause the gradients to become too small or too large, leading to unstable training and degraded performance. The gating mechanism of both \ac{gru} and \ac{lstm} models addresses this issue.} we opt for a \ac{gru} as \ac{rnn} for the encoder. The main steps of the \ac{gru} are
\begin{equation}
\label{eq: gru}
\begin{gathered}
        \vz_t = \sigma(\vW_z \mathrm{concat}(\tilde{\vh}_t, \vu_{t-1}) + \vb_z)\\
    \vr_t = \sigma(\vW_r \mathrm{concat}(\tilde{\vh}_t, \vu_{t-1}) + \vb_r)\\
    \tilde{\vu}_t  =\tanh(\vW_{\tilde{u}} \mathrm{concat}(\tilde{\vh}_t, \vr_t\odot\vu_{t-1}) + \vb_{\tilde{u}})\\
    \vu_t = (\ones-\vz_t) \odot\tilde{\vu}_t + \vz_t \odot \vu_{t-1},
\end{gathered}
\end{equation}
where $\vz_t$ and $\vr_t$ represent the update and the reset gate, respectively, and $\sigma$ denotes the sigmoid activation function~\cite{eslii}.
In particular, when $\vz_t$ is close to $\boldsymbol{1}$, we ignore completely the current input $\tilde{\vh}_t$ for the update of the current hidden state $\vu_t$. On the other hand, when both $\vr_t$ and $\vz_t$ are equal to zero, the hidden state only depends on the current input. 
The decoder also comprises a \ac{gru}. However, at each step, and in order to encourage the decoder to leverage the important parts of the encoder outputs before making the prediction, an attention mechanism with respect to the encoder outputs precedes the \ac{gru}. 
The model used for the decoder is shown in Fig.~\ref{fig:seq2seq}. In particular, the current hidden state, and the current input are concatenated, and then projected via a single layer onto a dimension equal to the maximum number of encoder outputs $\ell_{\max}$. At this point, the first $\ell$ out of the $\ell_{\max}$ units are selected and the obtained vector is normalized with the $\softmax$ activation function~\cite{eslii} to obtain the weights (probabilities), which are multiplied with the reversed encoder outputs $\vu_{\ell}, \dots, \vu_{1}$. 
\cb{The rationale behind reversing the encoder outputs is similar to the idea of using the reverse positional encoding in the transformer encoder. Essentially, by reversing the encoder outputs, we ensure that the weights associated with the initial units out of the $\ell_{\max}$ units correspond to the most recent known slots. This enables the network to generalize to sequences of varying lengths. Alternatively, instead of reversing the encoder outputs, we could achieve the same goal by selecting the last (instead of the first) $\ell$ units out of $\ell_{\max}$ before applying the $\mathrm{softmax}$.}
\\
Next, the result of the weighted sum of the encoder outputs, or ``attention'' with respect to the encoder outputs, is combined with the current decoder input and fed to a linear layer followed by a rectified linear unit (ReLU)~\cite{Goodfellow-et-al-2016} activation function to produce the second input vector for the decoder \ac{gru}. 
The current hidden state of the decoder \ac{gru} serves as the first input vector. Therefore, before entering the \ac{gru}, the current decoder input is preprocessed to take into account the contribution of the known slots.\\
Finally, and analogously with typical \acp{rnn}, the output of the \ac{gru} is fed to a linear layer to output the prediction of the next \ac{csi} vector. Like the Transformer-RPE model, the first input of the decoder is represented by the last known \ac{csi} snapshot, and teacher-forcing~\cite{WilliamsZ89} is deployed during training. However, differently from the Transformer-RPE model, in the \ac{seqa}-R model, the training happens sequentially.

\section{Benchmarks}
\label{sec:benchmarks}
In this work, we consider an \ac{lstm}~\cite{lstm-original-paper} model as further benchmark.
An \ac{lstm} cell employs three different gates, an input gate $\vi_t$, a forget gate $\vf_t$, and an output gate $\vo_t$ to prevent exploding or vanishing gradients. The main idea behind the gating system is not to retain information about all the inputs and to capture long-term dependencies. In formulas, we have:
\begin{equation}
\begin{gathered}
    \label{eq:lstm}
    \vi_t = \sigma(\vW_i \mathrm{concat}(\tilde{\vh}_t, \vu_{t-1}) + \vb_i)\\
    \vf_t = \sigma(\vW_f \mathrm{concat}(\tilde{\vh}_t, \vu_{t-1}) + \vb_f)\\
    \vo_t = \sigma(\vW_o \mathrm{concat}(\tilde{\vh}_t, \vu_{t-1}) + \vb_o)\\
    \tilde{\vc}_t  =\tanh(\vW_{\tilde{c}} \mathrm{concat}(\tilde{\vh}_t, \vu_{t-1}) + \vb_{\tilde{c}})\\
    \vc_t = \vf_t \odot \vc_{t-1} + \vi_t \odot \tilde{\vc}_t, \qquad
    \vu_t = \vo_t \odot \tanh{(\vc_t)},
\end{gathered}
\end{equation}
where $\vu_t$ and $\vc_t$ denote the hidden and the cell states, respectively, and $\sigma$ denotes the sigmoid activation function~\cite{eslii}. Note that differently from the classical \ac{rnn}, where the next hidden state is directly represented by $\tilde{\vc}_t$, here, the hidden state is updated using the cell state $\vc_t$. Therefore, $\tilde{\vc}_t$ is first modulated by the input gate and then by the output gate.
In this work, we utilize an \ac{lstm} to encode the input sequence $\{\tilde{\vh}_{i}\}_{i=1}^{\ell}$ into $\vu_{\ell}$. Then, in order to predict the next $\delta$ \ac{csi} slots, we employ a final linear layer that takes  $\vu_{\ell}$ as input and directly outputs $\vy = [\tilde{\vh}_{\ell + 1}^{\Tr}, \dots, \tilde{\vh}_{\ell + \delta}^{\Tr}]^{\Tr}$.

Apart from the \ac{lstm}-based model, additional benchmarks are: 
\textit{i)} a two-layer \ac{mlp} with a ReLU~\cite{Goodfellow-et-al-2016} activation function in the hidden layer;
\textit{ii)} a \ac{mar} model of order equal to $\ell$, where the coefficients are found with the ordinary least squares solution, see~\cite[Section 3.4.3]{weisberg};
\textit{iii)} a transformer which utilizes the standard \ac{pe} in the encoder, as in~\cite{VaswaniSPUJGKP17};
\textit{iv)} the Transformer-Parallel architecture, proposed in~\cite{transformer-parallel}.

\section{Simulations}
\label{sec:sims}
\subsection{Simulation setup}
For the simulations, we consider \ac{csi} sequences generated with QuaDRiGa v.2.6, see~\cite{quadriga}. In particular, we generate $N_{\text{samples}} =$ 150,000 \ac{csi} sequences corresponding to 1,500 different velocities. Therefore, we have $100$ users for each velocity. Every \ac{csi} sequence corresponds to a frame which contains $N_{\text{slot}}=20$ slots, each with duration $T_{\text{slot}}=0.5$\,ms. The carrier frequency is $2.6$\,GHz and each velocity $v$ measured in m/s is Rayleigh distributed:
$\frac{v}{\gamma^2}\exp{(-\frac{v^2}{2\gamma^2})}$, where $\gamma=8$. The reason for this choice is to simulate a realistic urban scenario, where the majority of the MTs move within a range of $20$ to $50$~km/h. However, there are MTs with $v < 20$~km/h (e.g., pedestrians and cyclists), as well as a few MTs with velocities exceeding $100$~km/h (e.g., fast moving cars). The scenario is the ``\text{BERLIN\_UMa\_NLOS}'' which generates non-line-of-sight channels with $25$ paths. 
The \ac{bs} positioned at a height of $25$\,m is equipped with a uniform rectangular array with $M=32$ antennas, with $8$ vertical and $4$ horizontal antenna elements. 
The users' initial positions are randomly distributed over a sector of $120\deg$, and with a minimum and maximum distance from the \ac{bs} of $50$\,m and $150$\,m, respectively, and at a height of $1.5$\,m.
All the generated \ac{csi} sequences are first normalized by the path-gain, and subsequently, they are split into training, validation, and test set with a percentage of $80\%$, $10\%$, and $10\%$, respectively.
We consider different noise levels for our simulations. In particular, we corrupt the channel $\vh_i$ as described in Eq.~\eqref{eq:noisy-obs} according to a noise variance $\sigma_n^2$ which fulfills a certain average $\SNR$ level. Therefore, given the average $\SNR$ we can determine $\sigma_n^2$ using the formula
\begin{equation}
    \SNR = \frac{\frac{1}{N_{\text{samples}}N_{\text{slot}}}\sum_{j=1}^{N_{\text{samples}}} \sum_{i=1}^{N_{\text{slot}}} \norm{\vh_i^{(j)}}^2}{M\sigma_n^2}
\end{equation}
where $\vh_i^{(j)}$ denotes the \ac{csi} vector in the i'th slot of the j'th sample in the dataset.
For our simulations, we assume that the first $\ell=16$ noisy \ac{csi} realizations are known. Therefore, the goal is to predict the next $\delta=4$ noisy \ac{csi} vectors. 
The performance metric that we consider is the \ac{nmse} with respect to the test set between the true noiseless \ac{csi} and the \ac{csi} predicted by the different models based on the noisy observations of the previous slots. In formulas, we have:
\begin{equation}
    \NMSE = \frac{1}{N_{\text{test}}}\sum_{j=1}^{N_{\text{test}}}\epsilon_j^2\quad \epsilon_j = \frac{\norm{\vH^{(j)} - \hat{\vH}^{(j)}}_{\mathrm{F}}}{\norm{\vH^{(j)}}_{\mathrm{F}}}
\end{equation}
where $\vH$ is the matrix that consists of the clean \ac{csi} snapshots, and  $\hat{\vH} = [\hat{\vh}_{\ell +1}, \dots, \hat{\vh}_{\ell + \delta}]$ is the matrix that consists of the corresponding predicted \ac{csi} snapshots.
During training, we assume that only noisy data are available. Therefore, the loss function is given by the $\NMSE$ between $\tilde{\vH} = [\tilde{\vh}_{\ell +1}, \dots, \tilde{\vh}_{\ell + \delta}]$ and $\hat{\vH}$.
All the models are trained separately for each $\SNR$ value. The number of epochs is set to $500$ and the batch size is equal to $200$. The Adam optimizer (see~\cite{KingmaB14}) with learning rate equal to $10^{-3}$ is used. For each model, the parameters leading to the smallest $\NMSE$ with respect the validation set, between the noisy \ac{csi} and the predicted one, are saved and considered during the test phase.
Note that the clean \ac{csi} is only used during the testing phase to evaluate the performance, whereas during training and validation, only noisy \ac{csi} observations are used.

\subsection{Models' parameters}

For the Transformer-RPE described in Section~\ref{sec:tnn} we consider $L_{\text{enc}}=L_{\text{dec}}=2$. Furthermore, in Algorithm~\ref{alg:mha} we set $H=4$, and  $d_{\text{attn}}=d_{\text{mid}}=16$, while the observed dimensions that correspond to the dimension of the real-valued \ac{csi} snapshots are $d_{\text{x}} = d_{\text{y}} = d_{\text{out}} = 64$. Consequently, $d_{\text{model}}=64$. The \ac{mlp} block in both, the encoder and the decoder, has a hidden dimension equal to $128$. 

For the \ac{seqa}-R model described in Section~\ref{sec:seq}, we consider \acp{gru} with two layers and hidden states with dimension equal to $128$. In the decoder, the first linear layer followed by the $\softmax$ activation maps $(64 + 128)$ units to $\ell_{\max} = 20$ units, where $64$ is the input dimension (the dimension of $\tilde{\vh}_i$), $128$ is the dimension of the hidden state $\vu_{i}$,  and the addition is due to the concatenation of the two. 
On the other hand, the second linear layer followed by a ReLU maps $(64 + 128)$ units to $64$ units, where $128$ is the resulting dimension after the ``Multiply \& Add'' block. The decoder's final linear layer, which produces the \ac{csi} prediction for the next step, maps the $128$ units of the next hidden state to the $64$ units of the next \ac{csi}. The dimensions of all the weights matrices and bias vectors of the \ac{gru}, that appear in Eq.~\eqref{eq: gru} can be derived with the given information.

For the \ac{lstm} model, we have considered a two-layer \ac{lstm}, with hidden states with dimension equal to $128$, and a final linear layer which maps the last hidden state to the output dimension which is equal to $256$. The dimensions of all the parameters which appear in Eq.~\eqref{eq:lstm} can be inferred with the given information. 
In the \ac{mlp} model, the observed dimension, the hidden dimension, and the output dimension are $1024$, $512$, and $256$, respectively.
The Transfomer with standard \ac{pe} has the same parameters as the Transformer-RPE.
For the Transformer-Parallel, the decoder input is initialized with the past 8 \ac{csi} snapshots followed by $\delta$ snapshots initialized as all-zeros vectors, while the remaining parameters have the same values as for the Transformer-RPE.

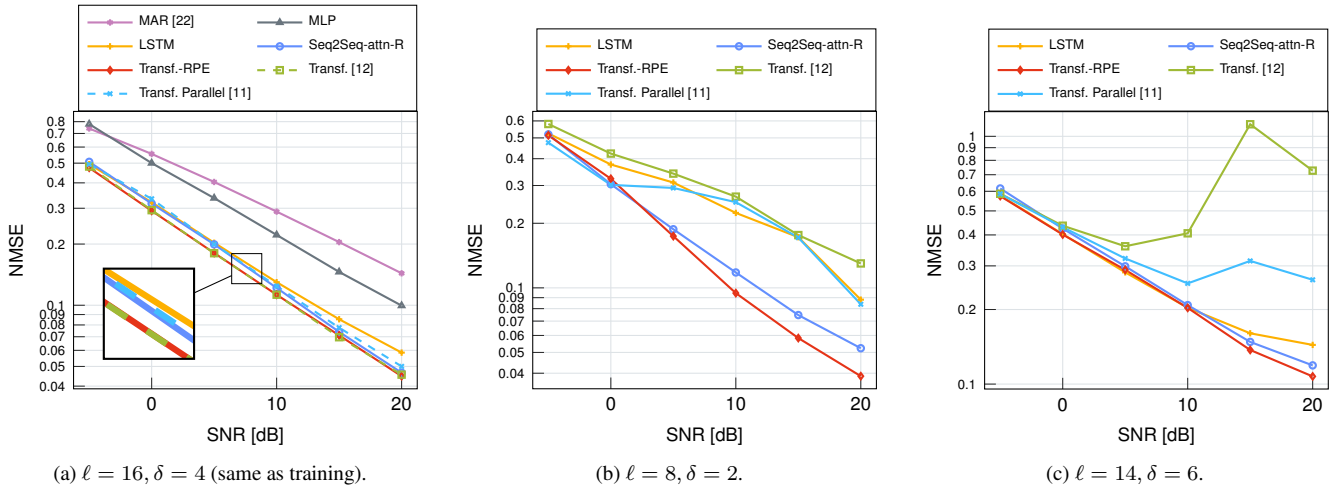
\begin{figure*}[t]
    \centering
  \subfloat[$\ell=16, \delta=4$ (same as training).\label{same}]{%
      \scalebox{1.0}{
\pgfplotsset{
  tick label style = {font=\sansmath\sffamily\scriptsize},
  every axis label = {font=\sansmath\sffamily\scriptsize},
  legend style = {font=\sansmath\sffamily\scriptsize},
  label style = {font=\sansmath\sffamily\scriptsize}
}

\begin{tikzpicture}[every axis plot/.append style={thick}, every axis/.append style={}, spy using outlines={rectangle, magnification=3, size=0.8cm, connect spies, }]

\definecolor{cornflowerblue99143254}{RGB}{99,143,254}
\definecolor{darkslategray38}{RGB}{38,38,38}
\definecolor{deeppink23654141}{RGB}{236,54,141}
\definecolor{gray}{RGB}{128,128,128}
\definecolor{indigo683129}{RGB}{68,3,129}
\definecolor{lightgray204}{RGB}{204,204,204}
\definecolor{lightpink255165165}{RGB}{255,165,165}
\definecolor{orange2531760}{RGB}{253,176,0}
\definecolor{turquoise81229255}{RGB}{81,229,255}
\definecolor{yellowgreen18121137}{RGB}{181,211,37}

\begin{axis}[
axis line style={black},
legend columns = 2,
legend style={
  font=\sansmath\sffamily\tiny,
  at={(1,1)},
  anchor=south east},
legend cell align={left},
log basis y={10},
tick align=inside,
tick pos=left,
width=175pt,
height=150pt,
x grid style={tum-grey-7},
xlabel={$\SNR$~[dB]},
xmin=-6.25, xmax=21.25,
xtick style={color=darkslategray38},
y grid style={tum-grey-7},
ylabel={$\NMSE$},
ymin=0.038810088990649, ymax=0.898223997573188,
ymode=log,
xmajorgrids,
ymajorgrids,
xminorgrids,
yminorgrids,
ytick style={color=darkslategray38},
ytick={0.001, 0.002, 0.003, 0.004, 0.005, 0.006, 0.007, 0.008, 0.009, 0.01,0.02, 0.03, 0.04, 0.05, 0.06, 0.07, 0.08, 0.09, 0.1, 0.2, 0.3, 0.4, 0.5, 0.6, 0.7, 0.8, 0.9, 1, 2, 3, 4, 5, 6, 7, 8, 9, 10},
yticklabels={
  \tiny {10^{-3}},\tiny ,\tiny ,\tiny ,\tiny ,\tiny ,\tiny ,\tiny ,\tiny ,\tiny {0.01},\tiny{0.02} ,\tiny{0.03} ,\tiny {0.04},\tiny{0.05} ,\tiny {0.06} ,\tiny{0.07} ,\tiny {0.08} ,\tiny{0.09} ,\tiny {0.1},\tiny{0.2} ,\tiny{0.3}  ,\tiny {0.4},\tiny {0.5},\tiny {0.6},\tiny {0.7},\tiny{0.8} ,\tiny{0.9} ,\tiny {1},\tiny ,\tiny ,\tiny ,\tiny ,\tiny ,\tiny ,\tiny ,\tiny , 
  \tiny {10^{1}},
}
]
\addplot [thick, tum-pink-1, mark=asterisk, mark size=1.2, mark options={solid,fill opacity=0}]
table {%
-5 0.740756392478943
0 0.55527800321579
5 0.404034048318863
10 0.288852721452713
15 0.204459860920906
20 0.143608421087265
};
\addlegendentry{MAR~\cite{weisberg}}
\addplot [thick, tum-grey-4, mark=triangle, mark size=1.2, mark options={solid,fill opacity=0}]
table {%
-5 0.778690099716187
0 0.500296831130981
5 0.336622774600983
10 0.221765980124474
15 0.145886391401291
20 0.0995517447590828
};
\addlegendentry{MLP}
\addplot [thick, orange2531760, mark=+, mark size=1.2, mark options={solid,fill opacity=0}]
table {%
-5 0.494563609361649
0 0.32092359662056
5 0.20293365418911
10 0.12987793982029
15 0.0853309258818626
20 0.0584767907857895
};
\addlegendentry{LSTM}
\addplot [thick, cornflowerblue99143254, mark=o, mark size=1.2, mark options={solid,fill opacity=0}]
table {%
-5 0.507529973983765
0 0.316383123397827
5 0.199346631765366
10 0.121726743876934
15 0.0739398449659348
20 0.046568725258112
};
\addlegendentry{\ac{seqa}-R}
\addplot [thick, TUMBeamerRed, mark=diamond, mark size=1.2, mark options={solid,fill opacity=0}]
table {%
-5 0.474680155515671
0 0.291470021009445
5 0.180774793028831
10 0.112653285264969
15 0.0711218267679214
20 0.0447676852345467
};
\addlegendentry{Transf.-RPE}
\addplot [thick, dashed, tum-green, mark=square, mark size=1.2, mark options={solid,fill opacity=0}]
table {%
-5 0.481210500001907
0 0.293589353561401
5 0.180316045880318
10 0.112694263458252
15 0.0696141272783279
20 0.0455651991069317
};
\addlegendentry{Transf.~\cite{VaswaniSPUJGKP17}}
\addplot [thick, dashed, TUMBeamerLightBlue, mark=x, mark size=1.2, mark options={solid,fill opacity=0}]
table {%
-5 0.493178635835648
0 0.333969265222549
5 0.199634864926338
10 0.124237142503262
15 0.0775928646326065
20 0.0500149317085743
};
\addlegendentry{Transf.~Parallel~\cite{transformer-parallel}}
\end{axis}
\spy[size=1.2cm,magnification=3] on (2.3,1.6) in node[fill=white] at (1.0, 1.0);
\end{tikzpicture}}}
      \hspace{0.3cm}
  \subfloat[$\ell=8, \delta=2$.\label{short}]{%
        \scalebox{1.0}{
\pgfplotsset{
  tick label style = {font=\sansmath\sffamily\scriptsize},
  every axis label = {font=\sansmath\sffamily\scriptsize},
  legend style = {font=\sansmath\sffamily\scriptsize},
  label style = {font=\sansmath\sffamily\scriptsize}
}

\begin{tikzpicture}[every axis plot/.append style={thick}, every axis/.append style={}]

\definecolor{cornflowerblue99143254}{RGB}{99,143,254}
\definecolor{darkslategray38}{RGB}{38,38,38}
\definecolor{deeppink23654141}{RGB}{236,54,141}
\definecolor{gray}{RGB}{128,128,128}
\definecolor{indigo683129}{RGB}{68,3,129}
\definecolor{lightgray204}{RGB}{204,204,204}
\definecolor{lightpink255165165}{RGB}{255,165,165}
\definecolor{orange2531760}{RGB}{253,176,0}
\definecolor{turquoise81229255}{RGB}{81,229,255}
\definecolor{yellowgreen18121137}{RGB}{181,211,37}

\begin{axis}[
axis line style={black},
legend columns = 2,
legend style={
  font=\sansmath\sffamily\tiny,
  at={(1,1)},
  anchor=south east},
legend cell align={left},
log basis y={10},
tick align=inside,
tick pos=left,
width=175pt,
height=150pt,
x grid style={tum-grey-7},
xlabel={$\SNR$~[dB]},
xmin=-6.25, xmax=21.25,
xtick style={color=darkslategray38},
y grid style={tum-grey-7},
ylabel={$\NMSE$},
xmajorgrids,
ymajorgrids,
xminorgrids,
yminorgrids,
ymin=0.0338560372109407, ymax=0.664604689069819,
ymode=log,
ytick style={color=darkslategray38},
ytick={0.001, 0.002, 0.003, 0.004, 0.005, 0.006, 0.007, 0.008, 0.009, 0.01,0.02, 0.03, 0.04, 0.05, 0.06, 0.07, 0.08, 0.09, 0.1, 0.2, 0.3, 0.4, 0.5, 0.6, 0.7, 0.8, 0.9, 1, 2, 3, 4, 5, 6, 7, 8, 9, 10},
yticklabels={
  \tiny {10^{-3}},\tiny ,\tiny ,\tiny ,\tiny ,\tiny ,\tiny ,\tiny ,\tiny ,\tiny {0.01},\tiny{0.02} ,\tiny{0.03} ,\tiny {0.04},\tiny{0.05} ,\tiny {0.06} ,\tiny{0.07} ,\tiny {0.08} ,\tiny{0.09} ,\tiny {0.1},\tiny{0.2} ,\tiny{0.3}  ,\tiny {0.4},\tiny {0.5},\tiny {0.6},\tiny {0.7},\tiny{0.8} ,\tiny{0.9} ,\tiny {1},\tiny ,\tiny ,\tiny ,\tiny ,\tiny ,\tiny ,\tiny ,\tiny , 
  \tiny {10^{1}},
}
]
\addplot [thick, orange2531760, mark=+, mark size=1.2, mark options={solid,fill opacity=0}]
table {%
-5 0.524799704551697
0 0.375773102045059
5 0.309719651937485
10 0.223563060164452
15 0.17223559319973
20 0.0881021991372108
};
\addlegendentry{LSTM}
\addplot [thick, cornflowerblue99143254, mark=o, mark size=1.2, mark options={solid,fill opacity=0}]
table {%
-5 0.519016802310944
0 0.304083943367004
5 0.187628179788589
10 0.117971822619438
15 0.0746826231479645
20 0.0523084327578545
};
\addlegendentry{\ac{seqa}-R}
\addplot [thick, TUMBeamerRed, mark=diamond, mark size=1.2, mark options={solid,fill opacity=0}]
table {%
-5 0.513785064220428
0 0.322458058595657
5 0.174194678664207
10 0.0945249199867249
15 0.0583348721265793
20 0.0387619435787201
};
\addlegendentry{Transf.-RPE}
\addplot [thick, tum-green, mark=square, mark size=1.2, mark options={solid,fill opacity=0}]
table {%
-5 0.580488979816437
0 0.42259755730629
5 0.340840369462967
10 0.266163796186447
15 0.17621922492981
20 0.129958033561707
};
\addlegendentry{Transf.~\cite{VaswaniSPUJGKP17}}
\addplot[thick, TUMBeamerLightBlue, mark=x, mark size=1.2, mark options={solid,fill opacity=0}]
table {%
-5 0.475215822458267
0 0.301960647106171
5 0.292240887880325
10 0.250871419906616
15 0.17153063416481
20 0.0838526412844658
};
\addlegendentry{Transf. Parallel~\cite{transformer-parallel}}
\end{axis}

\end{tikzpicture}}}
        \hspace{0.3cm}
    \subfloat[$\ell=14, \delta=6$.\label{long}]{%
        \scalebox{1.0}{
\pgfplotsset{
  tick label style = {font=\sansmath\sffamily\scriptsize},
  every axis label = {font=\sansmath\sffamily\scriptsize},
  legend style = {font=\sansmath\sffamily\scriptsize},
  label style = {font=\sansmath\sffamily\scriptsize}
}

\begin{tikzpicture}[every axis plot/.append style={thick}, every axis/.append style={}]

\definecolor{cornflowerblue99143254}{RGB}{99,143,254}
\definecolor{darkslategray38}{RGB}{38,38,38}
\definecolor{deeppink23654141}{RGB}{236,54,141}
\definecolor{gray}{RGB}{128,128,128}
\definecolor{indigo683129}{RGB}{68,3,129}
\definecolor{lightgray204}{RGB}{204,204,204}
\definecolor{lightpink255165165}{RGB}{255,165,165}
\definecolor{orange2531760}{RGB}{253,176,0}
\definecolor{turquoise81229255}{RGB}{81,229,255}
\definecolor{yellowgreen18121137}{RGB}{181,211,37}

\begin{axis}[
axis line style={black},
legend columns = 2,
legend style={
  font=\sansmath\sffamily\tiny,
  at={(1,1)},
  anchor=south east},
legend cell align={left},
log basis y={10},
tick align=inside,
tick pos=left,
width=175pt,
height=150pt,
x grid style={tum-grey-7},
xlabel={$\SNR$~[dB]},
xmajorgrids,
ymajorgrids,
yminorgrids,
xmin=-6.25, xmax=21.25,
xtick style={color=darkslategray38},
y grid style={tum-grey-7},
ylabel={$\NMSE$},
ymin=0.0955183998614187, ymax=1.25780167532647,
ymode=log,
ytick style={color=darkslategray38},
ytick={0.001, 0.002, 0.003, 0.004, 0.005, 0.006, 0.007, 0.008, 0.009, 0.01,0.02, 0.03, 0.04, 0.05, 0.06, 0.07, 0.08, 0.09, 0.1, 0.2, 0.3, 0.4, 0.5, 0.6, 0.7, 0.8, 0.9, 1, 2, 3, 4, 5, 6, 7, 8, 9, 10},
yticklabels={
  \tiny {10^{-3}},\tiny ,\tiny ,\tiny ,\tiny ,\tiny ,\tiny ,\tiny ,\tiny ,\tiny {0.01},\tiny ,\tiny ,\tiny ,\tiny ,\tiny ,\tiny ,\tiny ,\tiny ,   \tiny{0.1},  \tiny {0.2},\tiny{0.3} ,\tiny {0.4},\tiny{0.5} ,\tiny {0.6},\tiny {0.7},\tiny {0.8},\tiny {0.9}, \tiny {1},\tiny ,\tiny ,\tiny ,\tiny ,\tiny ,\tiny ,\tiny ,\tiny , 
  \tiny {10^{1}},
}
]
\addplot [thick, orange2531760, mark=+, mark size=1.2, mark options={solid,fill opacity=0}]
table {%
-5 0.5767782330513
0 0.403077483177185
5 0.282416373491287
10 0.203310966491699
15 0.16033248603344
20 0.143920749425888
};
\addlegendentry{LSTM}
\addplot [thick, cornflowerblue99143254, mark=o, mark size=1.2, mark options={solid,fill opacity=0}]
table {%
-5 0.616082549095154
0 0.424334526062012
5 0.298751324415207
10 0.208340167999268
15 0.147998407483101
20 0.118978828191757
};
\addlegendentry{\ac{seqa}-R}
\addplot [thick, TUMBeamerRed, mark=diamond, mark size=1.2, mark options={solid,fill opacity=0}]
table {%
-5 0.57397598028183
0 0.401409953832626
5 0.289045512676239
10 0.20301441848278
15 0.137298047542572
20 0.107392646372318
};
\addlegendentry{Transf.-RPE}
\addplot [thick, tum-green, mark=square, mark size=1.2, mark options={solid,fill opacity=0}]
table {%
-5 0.586363434791565
0 0.435720950365067
5 0.359977513551712
10 0.406055957078934
15 1.11872839927673
20 0.726959347724915
};
\addlegendentry{Transf.~\cite{VaswaniSPUJGKP17}}
\addplot [thick, TUMBeamerLightBlue, mark=x, mark size=1.2, mark options={solid,fill opacity=0}]
table {%
-5 0.585633397102356
0 0.429712951183319
5 0.321396678686142
10 0.254913181066513
15 0.314153283834457
20 0.263626933097839
};
\addlegendentry{Transf. Parallel~\cite{transformer-parallel}}
\end{axis}

\end{tikzpicture}}}
  \caption{$\NMSE$ vs. $\SNR$.}
  \label{fig:nmse-all} 
\end{figure*}

\subsection{Numerical results}
In Fig.~\ref{fig:nmse-all}, the $\NMSE$ vs. average $\SNR$ is displayed for all the models described within this work. 
In Fig.~\ref{same}, we can observe that the models designed for sequential data, such as \ac{lstm}, \ac{seqa}-R, and all the Transformer-based models, outperform both the \ac{mar} and the \ac{mlp} models. Moreover, among these, the models that include an attention mechanism outperform the \ac{lstm} model, despite the fact that the \ac{lstm} predict all the $\delta$ snapshots in one step. This means that in the \ac{lstm}, an imperfect prediction for the $\ell + 1$ snapshot has no influence regarding the prediction of the future $\delta - 1$ snapshots. 
Additionally, we observe that both the proposed Transformer-RPE and the one of~\cite{VaswaniSPUJGKP17} outperform the Transformer-Parallel~\cite{transformer-parallel} model in all the cases. This is because the former architectures can leverage the predicted \ac{csi} snapshots to make better prediction for the next one. And, at moderate or high $\SNR$ levels, this is advantageous for making a more accurate channel prediction.
\\
As a further benchmark, we observe how the sequential models that were trained for $\ell=16$ and $\delta=4$ perform when applied to sequence lengths different from the ones they were trained on.
In Fig.~\ref{short}, the $\NMSE$ of the different models for $\ell=8$ and $\delta=2$ is shown. We can see that the proposed Transformer-RPE and \ac{seqa}-R models considerably outperform all the other models, except for the low $\SNR$ cases, where the Transformer-Parallel is slightly better. However, in such low $\SNR$ cases, and for all the models, the $\NMSE$ is too high ($> 0.2$), which highlights that in this region all the models perform poorly.
\cb{Additionally, when comparing the results of the Transformer-RPE model and the Transformer of~\cite{VaswaniSPUJGKP17}, we can appreciate the improvements in terms of generalization capabilities that has been introduced with the RPE.} Note that, in Fig.~\ref{short} for the \ac{lstm}, only half of the vector $\vy$ described in Section~\ref{sec:benchmarks} is considered.
In Fig.~\ref{long}, the $\NMSE$ of the different models for $\ell=14$ and $\delta=6$ is shown. In this case, the proposed models outperform both, the Transformer with PE and the Transfomer-Parallel. At the same time, in Fig.~\ref{long}, we can observe that the performance of the \ac{lstm} is very close to those of both, the Transformer-RPE and the \ac{seqa}-R model. However, for the case in Fig.~\ref{long}, the \ac{lstm} model is included twice: the first time to predict the next 4 snapshots and the second time, it uses the known snapshots together with the predicted ones to predict the remaining 2 snapshots. In both, Fig.~\ref{short} and~\ref{long}, the presence of peaks in the lines of both, the Transformer with PE and the Transfomer-Parallel, is due to the fact that these models trained for $\ell=16$ and $\delta=4$ cannot generalize to other sequence lengths. \cb{These results highlight the robustness of the proposed models and show that they can generalize to any sequence length, as opposed to existing methods.}

\begin{table}[t]
    \centering
    \begin{tabular}{c|c|c}
         Model & \# parameters & \# FLOPs \\
         \hline & \\[-1.9ex]
         LSTM & $264,448$ & $6.37\times 10^6$\\
         \ac{seqa}-R & $370,832$ & $6.13\times 10^6$\\
         Transf.-RPE\footnotemark[3]  & $178,752$ & $5.96\times 10^6$\\
         Transf. Parallel~\cite{transformer-parallel} & $178,752$ & $4.68\times 10^6$
    \end{tabular}
    \caption{Number of parameters and complexity for $\ell=16$, $\delta=4$.}
    \label{tab:complexity}
\end{table}
\footnotetext[3]{same as Transf.~\cite{VaswaniSPUJGKP17}.}
In Table~\ref{tab:complexity}, we display the complexity of the different models both in terms of number of parameters and in terms of number of \acp{flop} for the case in which $\ell=16$ and $\delta=4$. We can observe that all the models designed for sequential data have a similar number of parameters and \acp{flop}. However, the transformer-based models require the smallest number of parameters, while the Transformer Parallel requires the smallest number of \acp{flop}. This is because the Transformer Parallel predicts all the $\delta$ \ac{csi} snapshots simultaneously, while the proposed Transformer-RPE uses sequential prediction that considers the contribution of previously predicted snapshots, and iterates over the transformer decoder $\delta$ times.

\cb{In summary, the proposed models, particularly the Transformer-RPE, offer more accurate channel prediction compared to existing models while maintaining the complexity at the same order of magnitude. Additionally, the fact that the proposed models exhibit robust results for different sequence lengths highlights that in a practical scenario it is sufficient to train a single model instead of having to train a different model for each combination of $\ell$ and $\delta$, which saves computational power, as well as storage requirements at the \ac{bs}.}

\section{Conclusions}
\label{sec:concl}
In this study, we introduce two models for channel prediction: Transformer-RPE and \ac{seqa}-R. Both models outperform existing methods in terms of channel prediction accuracy across various noise levels and can generalize to sequence lengths not encountered during training.
For future work, the proposed models can be extended to multiple-input-multiple-output (MIMO) channels, as opposed to just multiple-input-single-output (MISO) channels. Additionally, the models can be further developed to account for slots with varying durations.

\bibliographystyle{IEEEbib}
\bibliography{refs}

\end{document}